\documentclass[sigconf]{acmart}

\usepackage{multirow}

\AtBeginDocument{%
  }

\copyrightyear{2023}
\acmYear{2023}
\setcopyright{acmlicensed}\acmConference[MM '23]{Proceedings of the 31st ACM International Conference on Multimedia}{October 29-November 3, 2023}{Ottawa, ON, Canada}
\acmBooktitle{Proceedings of the 31st ACM International Conference on Multimedia (MM '23), October 29-November 3, 2023, Ottawa, ON, Canada}
\acmPrice{15.00}
\acmDOI{10.1145/3581783.3612458}
\acmISBN{979-8-4007-0108-5/23/10}

\acmSubmissionID{3303}

\begin{document}

\title{Local Consensus Enhanced Siamese Network with Reciprocal Loss for Two-view Correspondence Learning}
\author{Linbo Wang}{
\email{wanglb@ahu.edu.cn}
}

\author{Jing Wu}{
\email{e21301266@stu.ahu.edu.cn}
}

\author{Xianyong Fang}{
\authornote{Corresponding author.}
}

\author{Zhengyi Liu}{
\email{{fangxianyong, liuzywen}@ahu.edu.cn}
\affiliation{%
  \institution{School of Computer Science and Technology, Anhui University}
  \streetaddress{Jiulong Road \#111}
  \city{Hefei}
  \country{China}}
}

\author{Chenjie Cao}{
\email{ccjdurandal422@163.com}
\affiliation{%
  \institution{School of Data Science, Fudan University}
  \city{Shanghai}
  \country{China}}
}

\author{Yanwei Fu}{
\email{yanweifu@fudan.edu.cn}
\affiliation{%
  \institution{School of Data Science and Shanghai Key Lab of Intelligent Information Processing, Fudan University}
  \city{Shanghai}
  \country{China}}
\affiliation{%
  \institution{Fudan ISTBI-ZJNU Algorithm Centre for Brain-inspired Intelligence, Zhejiang Normal University}
  \city{Jinhua}
  \country{China}}
}


\begin{abstract}
Recent studies of two-view correspondence learning usually establish an end-to-end network to jointly predict correspondence reliability and relative pose. We improve such a framework from two aspects. First, we propose a Local Feature Consensus (LFC) plugin block to augment the features of existing models. Given a correspondence feature, the block augments its neighboring features with mutual neighborhood consensus and aggregates them to produce an enhanced feature. As inliers obey a uniform cross-view transformation and share more consistent learned features than outliers, feature consensus strengthens inlier correlation and suppresses outlier distraction, which makes output features more discriminative for classifying inliers/outliers. Second, existing approaches supervise network training with the ground truth correspondences
and essential matrix projecting one image to the other for an input image pair, without considering the information from the reverse mapping. We extend existing models to a Siamese network with a reciprocal loss that exploits the supervision of mutual projection, which considerably promotes the matching performance without introducing additional model parameters. Building upon MSA-Net \cite{zheng2022msa}, we implement the two proposals and experimentally achieve state-of-the-art performance on benchmark datasets.
\end{abstract}



\maketitle

\section{Introduction}
Discovering reliable feature correspondences has played a key role in many computer vision tasks, e.g., virtual reality \cite{szeliski1994image}, simultaneous location and mapping \cite{mur2015orb}, structure from motion \cite{schonberger2016structure}, image stitching \cite{brown2007automatic}, etc. Typically, the task is addressed in three steps, extracting local feature key points and descriptors using off-the-shelf detectors and descriptors, gathering putative correspondences by nearest neighbor searching in the descriptor space, and finally performing inlier identification for correspondence candidates. While the former two steps form the basis for the matching task, our focus here is the third step for high-quality feature correspondences. 

Putative correspondences generally contain a lot of mismatches, making the inlier/outlier classification a very challenging task. While traditional approaches~\cite{fischler1981random,ma2019locality,bian2017gms} have shown promising performances in limited scenes, recent studies usually explore a learning-based deep convolutional network for modeling the matching process in a data-driven manner. 

\begin{figure}[!tb]
\centering
\includegraphics[width=.48\textwidth]{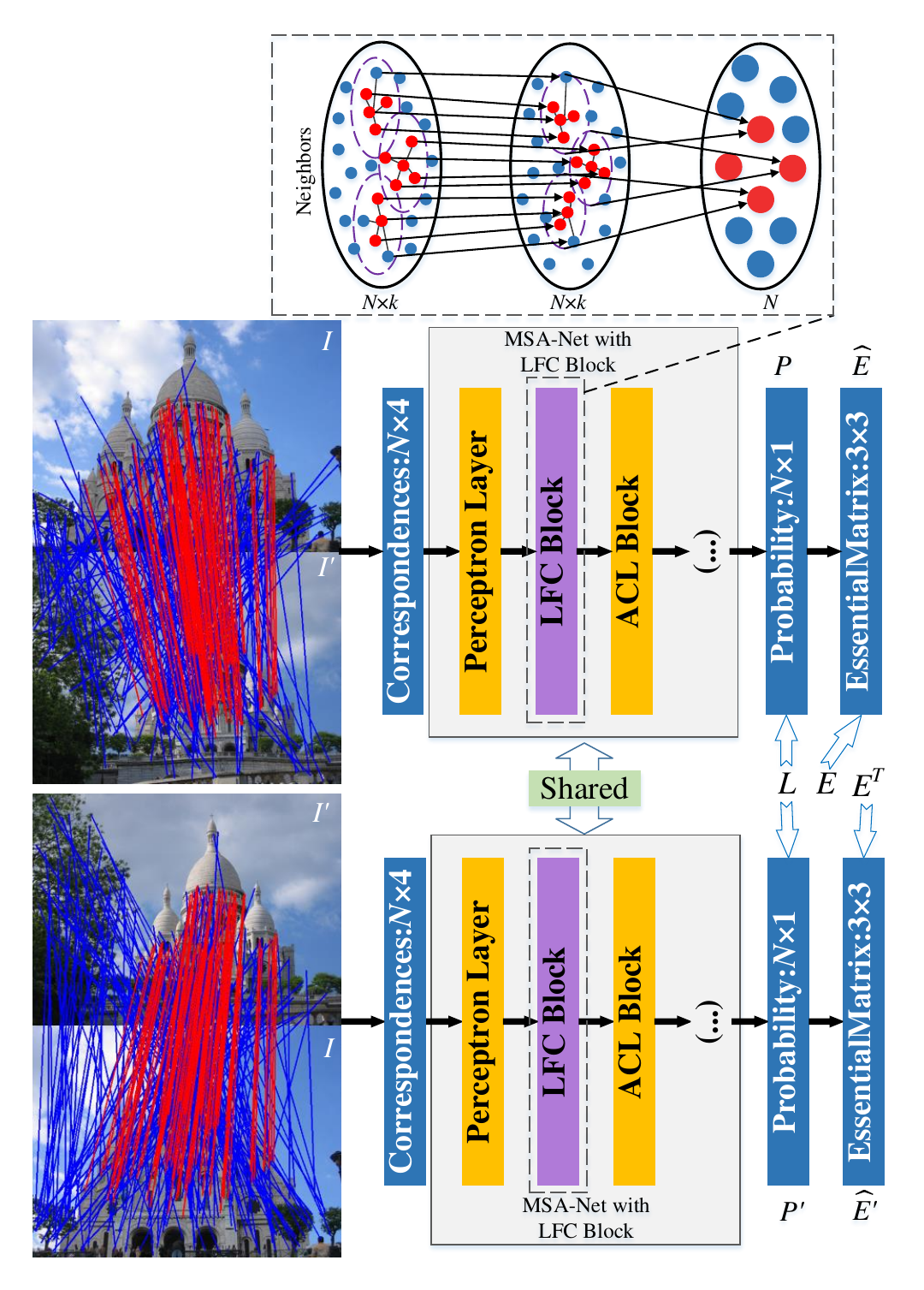}
\caption{The pipeline of MSA-Net \cite{zheng2022msa} extended Siamese network with the local feature consensus block. Two bidirectional putative correspondence sets are fed into a shared-weight network and two correspondences probability set $P$ and $P'$, and essential matrices $\hat{E}$ and $\hat{E'}$ are output. This process is jointly supervised by the ground truth label set $L$ and essential matrix $E$ with a reciprocal loss. Besides, a two-step local features consensus block is plugged inside to aggregate features of $k$ nearest neighbors for feature augmentation.}
\label{fig:pipeline}
\end{figure}

From the pioneering work of \cite{yi2018learning}, correspondence learning usually takes pairs of keypoint coordinates of putative matches as input, extracts feature maps with various convolutional blocks, and outputs the correspondence correctness and cross-view essential matrix. The two outputs are supervised by corresponding ground truth so that discriminative geometric features are learned to separate inliers/outliers. The framework has been extensively studied and achieved promising results. However, we argue that it can be further augmented from two aspects. As shown in Fig. \ref{fig:pipeline}, firstly, a local feature consensus block can be injected into existing models to boost features for inlier/outlier classification more effectively. Secondly, in contrast to the model only supervised by ground truth correspondences and epipolar constraint projecting one image to the other for an image pair, we demonstrate that information of mutual projection from each other can better guide the training process and enhance the matching performance. 

Local feature consensus is very helpful for distinguishing match inliers and outliers, which has been verified by many traditional methods\cite{wang2014progressive,wang2015common,bian2017gms,ma2019locality}. For two-view correspondence learning, inlier matches obey a uniform cross-view transformation but outliers do not. It indicates that geometric features of inliers extracted by an ideal matching network should be close to each other and quite different from those of outliers. However, this does not hold in practice due to the insufficient robustness of the network. To alleviate the issue, neighborhood feature aggregation is applied to obtain more consistent features among inliers. Basically, this
follows the idea of using neighboring features to vote for a better feature. Moreover, effective neighboring feature fusion requires to highlight the contribution of inlier neighbors while suppressing outlier distraction. This is achieved by neighboring feature boosting via mutual feature consensus given features of inliers are usually more similar than that of outliers. In particular, we adopt an attention-like process to measure the similarity of neighboring features and reconstruct each neighboring feature with a linear fusion of all neighboring ones based on similarity. 
Thereafter, all neighboring features are linearly fused with deformable attention-based weights, inspired by Deformable DETR \cite{zhu2020deformable}. We ensemble the two-step feature consensus into a feature augmentation plugin block, which can be conveniently integrated into existing networks. 

Proper supervision is vital for learning a robust matcher. Given a
pair of images $(I,I')$, existing approaches train networks by supervising
the prediction of correspondences $P$ and essential matrix
$\hat{E}$ projecting $I$ to $I'$ as shown in Fig. \ref{fig:pipeline}. Obviously,
this is one-way supervision, where no regularization is enforced on the reliability prediction $P'$ of reverse correspondences and the essential matrix $\hat{E}'$ projecting
$I'$ to $I$. Given the ground truth correspondence label $L$ and essential matrix $E$,
it is expected that $\hat{E}'=E^{T}$ and $P'=L$, which form natural constraints that can be used to regularize the matching network. However, existing matching models fail to take advantage of such regularization. To this end,
we extend the existing matching network to a Siamese one with a shared network structure as shown in Fig. \ref{fig:pipeline}. It first takes putative correspondences projecting $I$ to $I'$ and outputs $P$ and $\hat{E}$, and then takes reverse correspondences projecting $I'$ to $I$ and outputs $P'$ and $\hat{E}'$. A reciprocal loss is proposed to regularize $P$ and $P'$ to approximate $L$, $\hat{E}$ and $\hat{E'}$ to approximate $E$ and $E^T$ respectively. Consequently, the Siamese extension does not introduce any parameters over existing models, whilst considerably promoting their matching performance. 

The two proposals can be applied to a variety of existing approaches,
e.g., CNe \cite{yi2018learning}, OA-Net \cite{zhang2019learning}, MSA-Net \cite{zheng2022msa}, etc. 
Among them, we take MSA-Net as our baseline method in experiments.  
To summarize, our contributions are three-fold.

\begin{itemize}
\item A feature consensus plugin block boosts local deep features for more effective correspondence classification.  
\item A Siamese network is proposed by extending existing models with a shared network regularized with a reciprocal loss.
\item An improved MSA-Net equipped with the two proposals demonstrates superior performances over state-of-the-art competitors on standard datasets. 
\end{itemize} 

\section{Related Work}
\subsection{Parametric approaches} Traditionally, the two-view correspondence
task is addressed by generating a hypothetical projection model and
verifying its confidence. RANSAC \cite{fischler1981random} and its variants LO-RANSAC \cite{chum2003locally},
NG-RANSAC \cite{brachmann2019neural}, PROSAC \cite{chum2005matching} and USAC \cite{raguram2012usac} fall into this category. These approaches
usually sample a subset of correspondences to establish a parametric
model and evaluate its confidence by verifying how many correspondences
obey the model. This line of approaches performs well when most correspondence
outliers are removed in advance but may fail in robustness when the
outlier ratio is high.

\subsection{Non-parametric approaches} These approaches usually explore local
consensus for correspondence selection. Some studies \cite{wang2014progressive,wang2015common} project the correspondences
into the transformational space and identify inliers via neighborhood
density-based clustering. LPM \cite{ma2019locality} measures the structural inconsistency
around the two keypoints of a correspondence by counting mismatched neighboring
keypoint, thereby removing wrong matches. GMS \cite{bian2017gms}
applies grid-based motion consistency consensus to detect reliable
matches. Chen \textit{et al.} \cite{chen2013robust} define matches with similar local transformations
and keypoints located in the same local regions as neighbors and prunes
outliers of low neighborhood density. All these approaches make use
of neighborhood consensus for identifying match inliers and demonstrate
promising performances for handling non-rigid object matching and viewpoint
changes. However, constructing local neighborhoods in a heuristic manner
can be unreliable, and thus limits its application in specific scenes.
By contrast, we discover correspondence neighbors in deep
feature space and use local feature consensus to augment feature representation
in a data-driven manner. 

\subsection{Deep learning based approaches} CNe \cite{yi2018learning} is the
first to employ a convolutional neural network for inlier/outlier
match prediction. It adopts a PointNet-like architecture with context
normalization encoding global context in each correspondence, which
lays a good foundation for later research. OA-Net \cite{zhang2019learning} adds
a differentiable pooling layer to capture local context information
for improving model robustness. MSA-Net \cite{zheng2022msa} further integrates a multi-scale
attention module to mine local and global information for matching.
All these methods adopt dual-iterative networks to enhance
the performance, while T-Net \cite{zhong2021t} repeats the base network three
times and concatenates outputs of all sub-networks for final predictions.
It achieves superior performance at the cost of relatively more model
parameters. 

There are also studies adopting Transformer
to better model the process of correspondence prediction \cite{kipf2016semi,sun2021loftr}.
As to explore local consensus, NM-Net \cite{zhao2019nm} proposes a compatibility metric
to discover reliable neighbors and aggregate the neighboring features
with multiple convolution layers. CLNet \cite{zhao2021progressive} constructs a local-to-global dynamic graph to evaluate consensus scores for correspondences and
gradually remove outliers. MS$^2$DGNet \cite{dai2022ms2dg} builds sparse-semantics dynamic-graph network based on local neighborhood and employs a Transformer to
encode local structural information for performance enhancement. Akin
to MS$^2$DGNet, our approach also constructs a local neighborhood graph
for each correspondence but seeks to enhance neighboring features
with local consensus and further aggregate the feature with deformable
attention-based weights. More importantly, while existing approaches only use one-way
supervision for network optimization, we propose a reciprocal loss
and extend the existing network to a Siamese one for better correspondence
prediction. 

\section{Method}
\subsection{Problem Formulation}
Given a pair of two-view images $(I,I')$, the goal of two-view correspondence learning is to discover
their reliable matches and estimate the relative camera pose.
To this end, keypoints and corresponding descriptors are first extracted
by handcrafted or learning-based methods. Then, a putative
match matrix $C=[c_{1},c_{2},...,c_{N}]^{T}\in\mathbb{R}^{N\times4}$
is generated by nearest neighbor matching between descriptors, with
$c_{i}=[x_{i},y_{i},x_{i}',y_{i}']^{T}$ indicating keypoints
$(x_{i},y_{i})$ in $I$ and keypoints $(x_{i}',y_{i}')$ in $I'$
respectively. 

Typically, the task is cast as a joint problem of inlier/outlier match
classification and cross-view essential matrix estimation, which is
modeled by an end-to-end network with an iterative structure. Formally,
the process can be expressed as:
\begin{equation}
P_{t}=\begin{cases}
z_{\phi}(C), & t=1\\
z_{\varphi}([C||R_{t-1}||P_{t-1}]), & t=2
\end{cases}\label{eq:evalProb}
\end{equation}
\begin{equation}
R_{t}=h(C,\hat{E}_{t})\label{eq:evalResidual}
\end{equation}
\begin{equation}
\hat{E_{t}}=g(C,P_{t})\label{eq:evalEss}
\end{equation}
where $z_{\phi}(\cdot)$ and $z_{\varphi}(\cdot)$ are two sub-networks
with the same structure but different learnable parameters $\phi$
and $\varphi$; $[\cdot||\cdot]$ means feature concatenation; 
$P_{t}=[p_{1},p_{2},...,p_{N}]^T$ with $t$ indexing
the sub-networks and $p_{i}\in[0,1)$ indicating the correctness of
match $c_{i}$, e.g., $c_{i}$ is an outlier if $p_{i}=0$; $R_{t}$
and $\hat{E}_{t}$ represent the residual and essential matrix estimated
by the epipolar error function $h(\cdot,\cdot)$ and weighted eight-point
algorithm $g(\cdot,\cdot)$, respectively. Noticeably, the first sub-network
takes only $C$ as input, while its output $P_{1}$ and corresponding
residual $R_{1}$ are concatenated with $C$ to feed to the second
sub-network. The resultant $P_{2}$ and $\hat{E}_{2}$
are deemed as the final output.

\noindent\textbf{Loss Function}. Existing approaches usually adopt a combined loss for training the
model parameters $\phi$ and $\varphi$. Formally, it can be written
as
\begin{equation}
\small \mathcal{L}(C)=\sum_{t=1}^{2}\mathcal{L}_{t}(C),\;\mathcal{L}_{t}(C)=\mathcal{L}_{\text{cls}}(P_{t},L)+\lambda\mathcal{L}_{\text{reg}}(\hat{E}_{t},E)\label{eq:totalLoss}
\end{equation}
where $\mathcal{L}_{\text{cls}}$ is a binary cross entropy loss that
supervises the probability set $P_{t}$ to approximate the ground-truth
label set $L$; $\mathcal{L}_{\text{reg}}$ is a geometric loss to
penalize the difference between the estimated essential matrix $\hat{E}_{t}$
and ground-truth $E$; $\lambda$ is a hyper-parameter to balance the two losses. Specifically, $\mathcal{L}_{\text{reg}}$ is defined
over the inlier index set $\mathcal{I}$ by
\begin{equation}
\mathcal{L}_{\text{reg}}(\hat{E},E)=\sum_{i=1}^{|\mathcal{I}|}\frac{\left({\mathbf{x}'}^{T}_{\mathcal{I}_i}\hat{E}\mathbf{x}_{\mathcal{I}_i}\right)}{\left\Vert \rho\left(E\mathbf{x}_{\mathcal{I}_i}\right)\right\Vert+\left\Vert \rho\left(E^{T}\mathbf{x}'_{\mathcal{I}_i}\right)\right\Vert},
\label{eq:regLoss}
\end{equation}
where $\rho([a,b,c]^{T})=[a,b]^{T}$; $\mathcal{I}_{i}$ stands for the $i$-th index of inlier matches; $\mathbf{x}=[x,y,1]^T$ and $\mathbf{x}'=[x',y',1]^T$ represent two keypoints of a match.
\begin{figure}[!tb]
\centering
\includegraphics[width=.48\textwidth]{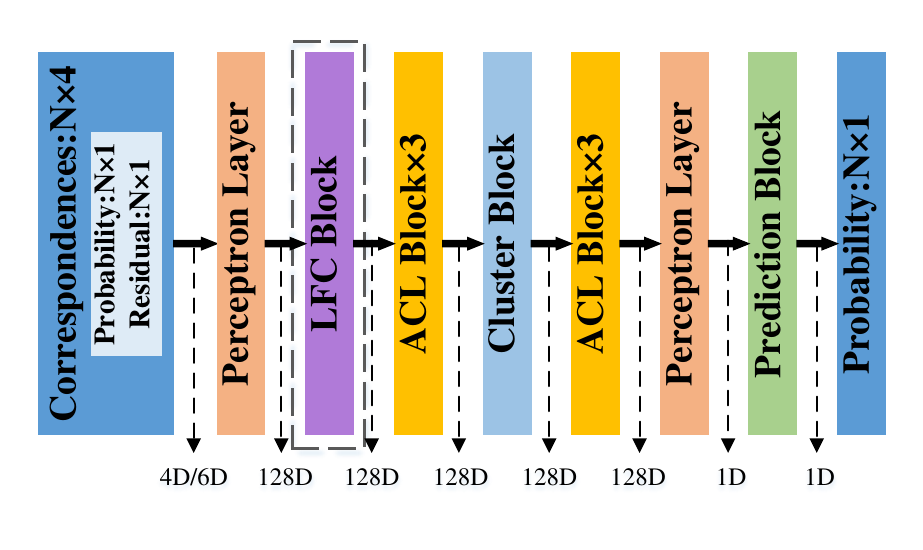}
\caption{Architecture of MSA-Net equipped with the local feature consensus block, which is marked by the dotted rectangle. Note that the MSA-Net consists of two similar sub-network with different parameters $z_{\phi}(\cdot)$ and $z_{\varphi}(\cdot)$. $z_{\phi}(\cdot)$ takes the putative correspondence matrix as input and outputs a matching probability vector and a residual vector, which are bundled with the putative correspondence matrix and fed to $z_{\varphi}(\cdot)$ for final correspondence prediction.}
\label{fig:msa-net}
\end{figure}

\subsection{Review of MSA-Net}
Fig. \ref{fig:msa-net} shows the base network $z(\cdot)$ of MSA-Net \cite{zheng2022msa}
with an LFC plugin block. MSA-Net consists
of a perceptron layer, three attentional correspondence learning (ACL) blocks, a cluster block, a multi-scale attention block, three
more ACL blocks, and a prediction block in order. The ACL block is
further composed of two context channel refinement blocks separated by
a multi-scale attention block, which follows the spirit of squeeze
and excitation to aggregate global and local features for robust matching.
The cluster block is proposed by OA-Net \cite{zhang2019learning}, while the prediction
block contains an MLP layer, a tanh, and a ReLU operation. We refer
the readers to \cite{zheng2022msa} for more details. 
MSA-Net performs well for two-view correspondence learning, however, it can be further augmented by two means, i.e., equipped with an LFC block and then extended to a Siamese network with a reciprocal loss. Next, we describe the two parts subsequently.

\subsection{Local Feature Consensus Block}
The LFC block aims at augmenting each matching feature by fusing nearby
ones. Before the fusion, we first
conduct a consensus-based feature boosting for neighbors of each match.
Aggregating the boosted neighboring features results in
more discriminative features facilitating inlier/outlier classification.

To this end, a local graph is constructed for each match $c_{i}$
as $\mathcal{G}_{i}=(\mathcal{V}_{i},\mathcal{E}_{i})$,
where $\mathcal{V}_{i}=\{c_{i}^{j}\}_{j=1,...,k}$ are $k$ neighbors
of $c_{i}$, and $\mathcal{E}_{i}$ is the edge set connecting $c_{i}$
and its neighbors $c_{i}^{j}$. Moreover, $c_{i}^{j}$ corresponds
to the $j$-th nearest neighbors of $c_{i}$ in feature space, measured
by Euclidean distance. Denote the features extracted for the match
set $C$ as $F=\{f_{i}\}_{i=1,...,N}$.
The edge feature can be written as \cite{wang2019dynamic,zhao2021progressive,dai2022ms2dg}
\begin{equation}
e_{i}^{j}=[f_{i}||f_{i}-f_{i}^{j}],j=1,...,k,\label{eq:edgeFeat}
\end{equation}
where $[\cdot||\cdot]$ means feature concatenation, and $f_{i}-f_{i}^{j}$
is the residual feature of $c_{i}$ and $c_{i}^{j}$.

Our goal then is to aggregate features of neighbors to construct a new feature $\{e_{i}^{j}\}_{j=1,...,k}\rightarrow\hat{f_{i}}$.
It can be fulfilled by two steps: 1) conducting neighboring feature consensus
via mapping $\{e_{i}^{j}\}_{j=1,..., N}\rightarrow\hat{e}_{i}^{j}$;
and 2) performing attention-based deformable feature fusion $\{\hat{e}_{i}^{j}\}_{j=1,..., N}\rightarrow\hat{f_{i}}$.
Next, we discuss the design of each step.
\begin{figure}[!tb]
\centering
\includegraphics[width=.48\textwidth]{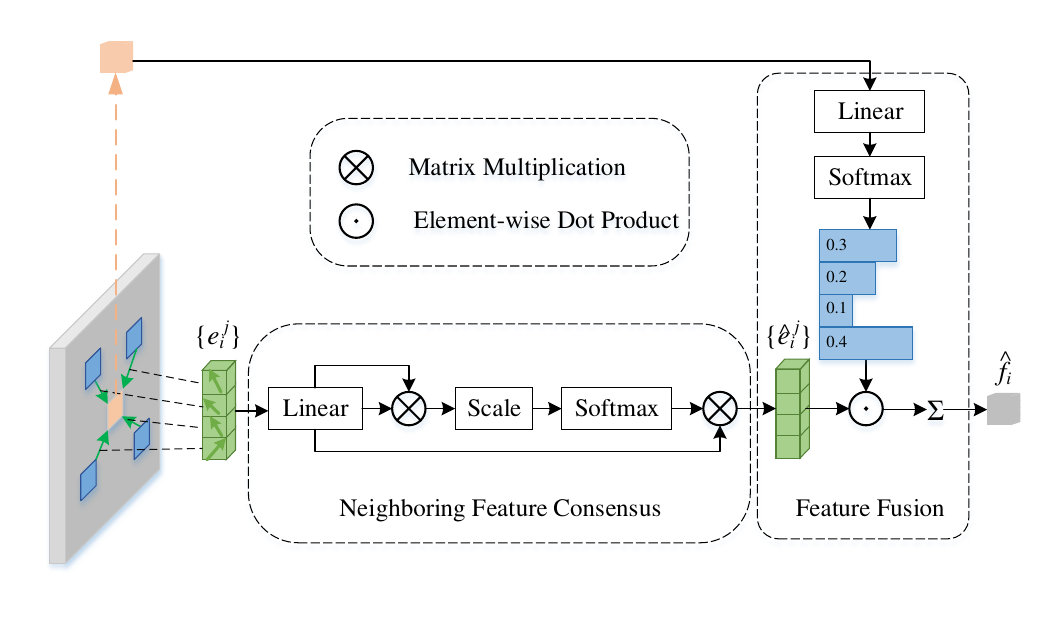}
\caption{Architecture of the local feature consensus block. Neighboring features are reconstructed by the mutual consensus with an attention-like operation and then aggregated via attention-based deformable weighting.}
\label{fig:consensus}
\end{figure}

\noindent\textbf{Neighboring Feature Consensus}. A naive way to
augment features is to adopt MLPs as in \cite{wang2019dynamic}. However, $1\times1$ kernels of MLPs mapping features separately in the spatial dimension and thus may discard context information \cite{zhao2021progressive}.
Therefore, we augment $e_{i}^{j}$ by exploring feature correlation-based mutual consensus.
Specifically, as shown in Fig. \ref{fig:consensus}, it performs an attention-like operation to
reconstruct the features. Formally, given the feature matrix $E_{i}\in\mathbb{R}^{k\times2d}$
by stacking $\{e_{i}^{j}\}_{j=1,...,k}$ along vertical dimension, the process can be written as
\begin{equation}
\hat{E}_{i}=\text{softmax}\left(\frac{\tilde{E_{i}}\tilde{E}_{i}^{T}}{\sqrt{d}}\right)\tilde{E_{i}}=A_{i}\tilde{E_{i}},\label{eq:selfAug}
\end{equation}
where $\tilde{E_{i}}=E_{i}W\in\mathbb{R}^{k\times d}$ is
a matrix projected by the learnable embedding matrix $W\in\mathbb{R}^{2d\times d}$,
which is shared by all correspondences. $A_{i}\in\mathbb{R}^{k\times k}$
encodes the feature correlation among all feature neighbors. 
As inlier neighbors obey a uniform cross-view transformation and share more consistent
learned features than outliers, mutual consensus can help strengthen 
feature correlation among inlier neighbors and alleviate outlier distraction during later fusion.
Besides, we also conduct multi-head augmentation
as self-attention does in practice. 

\begin{figure*}[!tb]
\centering
\includegraphics[width=.9\textwidth]{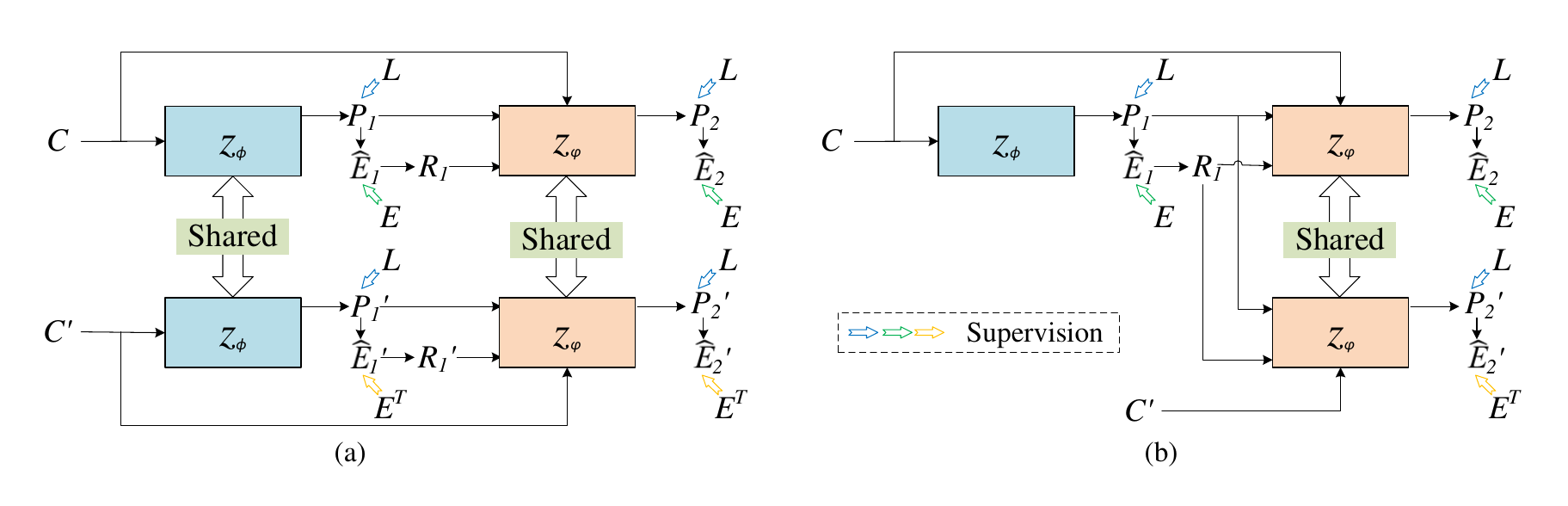}
\caption{Two designs of Siamese networks with different reciprocal loss. (a) Both $z_{\phi}(\cdot)$ and $z_{\varphi}(\cdot)$ are shared. (b) Only $z_{\varphi}(\cdot)$ is shared.}
\label{fig:siamese}
\end{figure*}
\noindent\textbf{Deformable Attention-based Feature Fusion}. Feature
fusion can be done by a linear-weighted features summation as
\begin{equation}
\hat{f}_{i}=\sum_{j=1}^{k}\omega_{i}^{j}\hat{e}_{i}^{j},\label{eq:linearFuse}
\end{equation}
where $\{\hat{e}_{i}^{j}\}_{j=1,...,k}$ are split from $\hat{E}_i$ in Eq. \ref{eq:selfAug}; the weight $\omega_{i}=[\omega_{i}^{1},...,\omega_{i}^{k}]^{T}$
is usually obtained by average pooling \cite{wang2019dynamic}, 
which fails to jointly consider features $f_i$ and $\hat{e}_{i}^{j}$ for evaluating $\hat{f}_{i}$.
Inspired by \cite{zhu2020deformable}, we set by $\omega_{i}$ with an attention based deformable weighting strategy. Formally, it can be expressed as
\begin{equation}
\omega_{i}=\text{softmax}(W'f_{i})\in\mathbb{R}^{k},\label{eq:evalWgt}
\end{equation}
where $W'\in\mathbb{R}^{k\times d}$ is a learnable linear projection matrix shared by features
of all matches. Note that deformable DETR \cite{zhu2020deformable} utilizes $\omega_{i}$ to weight neighboring features with learned spatial offsets, while we obtain $\hat{e}_{i}^{j}$ by the nearest neighbor searching in the feature space.
Despite the difference, our method still shares the same spirit with the deformation attention in
that it attends to a small set of key feature points for feature aggregation.

\subsection{Siamese Network with Reciprocal Loss}
MSA-Net takes the putative correspondence matrix $C$ as input and produces the matching probability set $P$
and the essential matrix $\hat{E}$. During training, the outputs are supervised by ground truth correspondence label $L$ and essential matrix $E$. Moreover, if we feed the reverse correspondence matrix $C'=[c_{1}',c_{2}',...,c_{N}']$ with $c'_{i}=[x_{i}',y_{i}',x_{i},y_{i}]$ to MSA-Net, it then outputs another probability set $P'$ and essential matrix $\hat{E}'$. Ideally, we have
\begin{equation}
P'=L\;\text{and}\;\hat{E}'=E^{T}.\label{eq:recipEq}
\end{equation}
However, the equations cannot be guaranteed as MSA-Net does not consider
any regularization on reverse projection, which can obviously promote
the robustness of the network if explored properly.
To this end, we extend the revised MSA-Net
to a Siamese network with a reciprocal matching loss for jointly supervising $P'$ and $P$ as well as $\hat{E}'$ and $\hat{E}^{T}$.

In particular, two Siamese networks with different loss functions are explored (Fig.~\ref{fig:siamese}).
The first Siamese network (Fig.~\ref{fig:siamese}a) consists of two branches sharing
both $z_{\phi}(\cdot)$ and $z_{\varphi}(\cdot)$. $C$ and $C'$ are fed to different branches to generate respective probability sets and essential matrices.
Thus, the loss function can be written as
\begin{equation}
\mathcal{L}_{\text{reciprocal}}^{(a)}=\mathcal{L}(C)+\mathcal{L}(C').\label{eq:recipLoss1}
\end{equation}
In the second Siamese network (Fig.~\ref{fig:siamese}b), $C'$ along with $P_{1}$ and $\hat{E}_{1}$
output by $z_{\phi}(\cdot)$ is directly fed to a shared module $z_{\varphi}(\cdot)$. Accordingly, the loss is defined as 
\begin{equation}
\mathcal{L}_{\text{reciprocal}}^{(b)}=\mathcal{L}(C)+\mathcal{L}_{2}(C'),\label{eq:recipLoss2}
\end{equation}
where ${L}_{2}(C')$ is the loss of $z_{\varphi}(C')$. Obviously, the first network optimized by $\mathcal{L}_{\text{reciprocal}}^{(a)}$ is computationally more expensive than
the second one. Interestingly, our experimental verification shows that the second network with $\mathcal{L}_{\text{reciprocal}}^{(b)}$ achieves comparable performance to the first one (see Sec. \ref{subsec:ablation}). This might be due
to the network $z_{\varphi}(\cdot)$ is more stable than $z_{\phi}(\cdot)$.
Therefore, in practice, we only use the second network for experiments.

Moreover, the proposed Siamese network and reciprocal loss have several merits. Firstly, it adopts shared weights without any new parameters and thus avoids degrading the training and testing efficiency. Secondly, it actually doubles the amount of training data for more robust network optimization, which can be viewed as one kind of data augmentation specialized for the task of feature matching. 
Thirdly, it uses ground-truth $L$ and $E^{T}$ to supervise the matching probability set and the essential matrix output of the reverse correspondence matrix $C'$, which implicitly regularizes the network to optimize the consistency between $P'$ and $P$ as well as $\hat{E}'$ and $\hat{E}^{T}$ and prominently boosts the matching robustness.
\begin{table*}[tbp]
  \centering
  \caption{Quantitative comparisons of correspondence prediction on the YFCC100M and SUN3D datasets under known and unknown scenes. The last two rows are our approaches without and with the Siamese extension. The bold font indicates the best result of each column.}
  \small
  \setlength{\tabcolsep}{1mm}{
    \begin{tabular}{c|ccc|ccc|ccc|ccc}
    \hline
    Datasets & \multicolumn{6}{c|}{YFCC100M(\%)}             & \multicolumn{6}{c}{SUN3D(\%)} \\
    \hline
    \multirow{2}{*}{Matcher} & \multicolumn{3}{c|}{Known Scene} & \multicolumn{3}{c|}{Unknown Scene} & \multicolumn{3}{c|}{Known Scene} & \multicolumn{3}{c}{Unknown Scene} \\
    \cline{2-13}
          & $P$     & $R$     & $F$     & $P$     & $R$     & $F$     & $P$     & $R$     & $F$     & $P$     & $R$     & $F$ \\
    \hline
    Point-Net++ \cite{qi2017pointnet++} & 49.62 & 86.19 & 62.98 & 46.39 & 84.17 & 59.81 & 52.89 & 86.25 & 65.57 & 46.30  & 82.72 & 59.37 \\
    \hline
    DFE \cite{ranftl2018deep}  & 56.72 & 87.16 & 68.72 & 54.00    & 85.56 & 66.21 & 53.96 & 87.23 & 66.68 & 46.18 & 84.01 & 59.60 \\
    \hline
    ACNe \cite{sun2020acne} & 60.02 & 88.99 & 71.69 & 55.62 & 85.47 & 67.39 & 54.11 & 88.46 & 67.15 & 46.16 & 84.01 & 59.58 \\
    \hline
    CNe \cite{yi2018learning}  & 54.43 & 86.88 & 66.93 & 52.84 & 85.68 & 65.37 & 53.70  & 87.03 & 66.42 & 46.11 & 83.92 & 59.37 \\
    \hline
    OA-Net++ \cite{zhang2019learning} & 60.03 & 89.31 & 71.80  & 55.78 & 85.93 & 67.65 & 54.30  & 88.54 & 67.32 & 46.15 & 84.36 & 59.66 \\
    \hline
    NM-Net \cite{zhao2019nm} & -     & -     & -     & 55.30  & 85.80  & 64.71 & -     & -     & -     & 46.68 & 83.98 & 56.34 \\
    \hline
    MS$^2$DG-Net \cite{dai2022ms2dg} & 63.17 & 90.98 & 74.57 & 59.11 & 88.40 & 70.85 & 54.50  & 88.63 & 67.50  & 46.95 & 84.55 & 60.37 \\
    \hline
    T-Net \cite{zhong2021t} & 62.14 & 91.70  & 74.08  & 57.48 & 88.39 & 69.66 & 54.98 & 88.82 & 67.92 & 46.94 & 84.53 & 60.36 \\
    \hline
    MSA-Net \cite{zheng2022msa} & 61.98 & 90.53 & 73.58  & 58.70  & 87.99 & 70.42  & 55.38 & 87.51 & 67.83  & 48.10  & 83.81 & 61.12  \\
    \hline
    Ours(MSA-LFC) & 64.38 & 91.85 & 75.70 & 59.67 & 88.42 & 71.25 & 55.82 & 88.78 & 68.54 & 47.86 & 84.84 & 61.20  \\
    \hline
    Ours  & \textbf{65.47 } & \textbf{91.94 } & \textbf{76.48 } & \textbf{60.84 } & \textbf{88.66 } & \textbf{72.16 } & \textbf{56.05 } & \textbf{88.93 }  & \textbf{68.76 } & \textbf{48.14 } & \textbf{85.09 } & \textbf{61.49 } \\
    \hline
    \end{tabular}
    }
  \label{tab:matchprd}%
\end{table*}%

\section{Experiments}
Two popular datasets, YFCC100 \cite{thomee2016yfcc100m} and SUN3D \cite{xiao2013sun3d},
are adopted in the experiments. Experimental results and comparisons are reported. We also present ablation studies on the YFCC dataset for testing various design choices of the proposed approach.

\subsection{Experimental Settings}
\textbf{Datasets}. YFCC100M introduced by Yahoo contains 100 million outdoor scene images, from which 72 sequences of different tourist landmarks are gathered \cite{heinly2015reconstructing}. Following OANet \cite{zhang2019learning}, 68 sequences are taken for training and the remaining 4 are treated as unknown scenes for testing all methods.
SUN3D is an indoor scene image dataset, originating from an RGB-D video dataset. 239 sequences are used for training while 15 are saved as unknown scenes for testing. Each sequence of image collection is constructed by sub-sampling every 10 frames from the original video. In addition, the training set of both datasets is further divided into three parts, training (60\%), validation (20\%), and testing (20\%). As the new test subset contains images of the same scene with the training set, it is termed as known scenes in comparison with the unknown scenes during experiments. 

\noindent\textbf{Evaluation Metrics}.
We follow the existing approach to set the evaluation metric for different tasks. Specifically, for the task of correspondence prediction, $Precision (P)$, $Recall (R)$, and $F$-$score$ are used to measure the performance. For the task of cross-view pose estimation, we evaluate the quality by mean Average Precision (mAP) under angular differences of 5$^\circ$ between ground truth and predicted vectors for rotation and translation.

\noindent\textbf{Competitors}.
We compare our method with several learning-based approaches, including  Point-Net++ \cite{qi2017pointnet++}, DFE \cite{ranftl2018deep}, ACNe \cite{sun2020acne}, CNe \cite{yi2018learning}, OA-Net++ \cite{zhang2019learning}, NM-Net \cite{zhao2019nm}, MS$^2$DG-Net \cite{dai2022ms2dg}, T-Net \cite{zhong2021t}, MSA-Net \cite{zheng2022msa}. For a fair comparison, all methods are trained and tested using the same setting on both datasets. Moreover, all competitors are tested with codes and models shared by respective authors. In case the trained model is not available, we train it using the shared code by ourselves. Besides, initial feature correspondences are established with SIFT features \cite{lowe2004distinctive}.

\noindent\textbf{Implementation Details}. We implement the proposed method with PyTorch and follow MSA-Net to set parameters. For training, the Adam optimizer is used with the learning rate 10$^{-3}$. The batch size is set to 32 and the network is trained for 500$k$ iterations. All experiments are conducted with a single NVIDIA GTX 4090 GPU. 
Moreover, in our implementation, LFC is empirically inserted after the first perception layer, even though it can be theoretically integrated after any block of MSA-Net. Besides, model parameters are learned using the Siamese structure. Once the training is over, LFC-injected MSA-Net is taken for testing. As a result, the testing efficiency keeps on par with that of MSA-Net.

\subsection{Correspondence Prediction}
Table \ref{tab:matchprd} shows the results of correspondence prediction. The proposed method outperforms all other ones in all metrics except that the highest recall rate for known scenes on the SUN3D dataset is reached by T-Net. For T-Net, it iterates the base networks three times and combines the output of all three sub-networks to produce the results with 3.78M parameters. Comparatively, OA-Net++, MS$^2$DG-Net, MSA-Net, and our method only repeats the base network 2 times with 2.47M, 2.61M, 1.62M, and 1.73M model parameters. Therefore, T-Net keeps a high recall rate at the cost of a bigger model. The proposed method exceeds T-Net in recall rate for most cases and with higher precision, producing the highest $F$-$score$ at all 4 scenes no matter with or without using the reciprocal loss. Moreover, in the absence of reciprocal loss, the proposed method obtains a considerable performance gain over MSA-Net in terms of all metrics in all four scenes, indicating the usefulness of augmenting features with local consensus. Besides, extending the local consensus-enhanced MSA-Net to the Siamese network can further boost the performance, suggesting the effectiveness of the proposed reciprocal loss. 

\begin{figure*}[!htb]
\centering
\includegraphics[width=.98\textwidth]{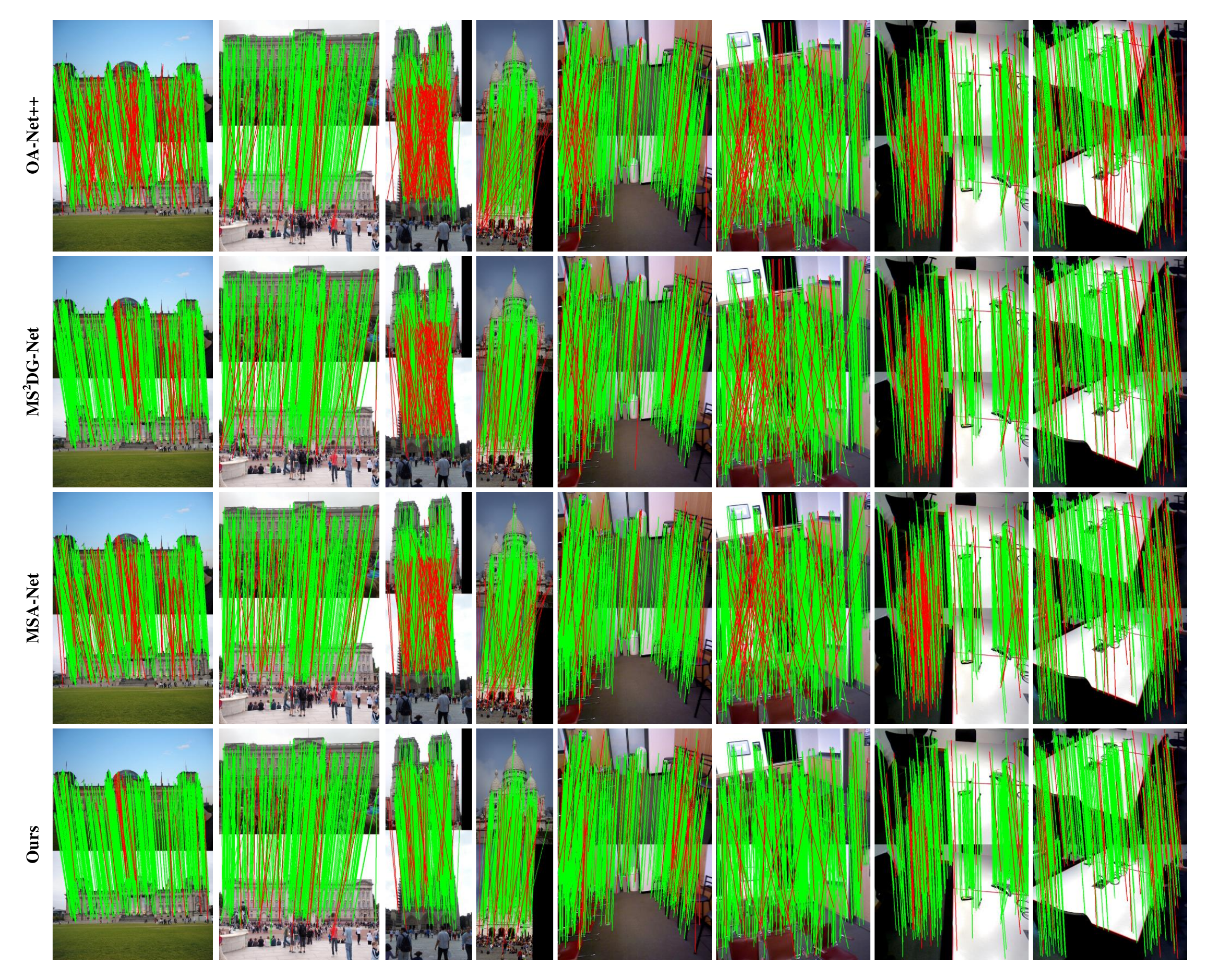}
\caption{Visualization results of OA-Net++, MS$^2$DG-Net, MSA-Net, and our proposed approach. Green and red lines indicate correct and incorrect correspondences, respectively.}
\label{fig:visual}
\end{figure*}

In Figure. \ref{fig:visual}, we present visual results of OA-Net, MS$^2$DG-Net, MSA-Net, and our approach without RANSAC-based post-processing. As shown, all inlier matches of an image pair obey a uniform global cross-image transformation, while outlier matches are in general randomly projected. As the matching networks all take only pairs of pixel coordinates of candidate matches as input to distinguish inliers/outliers, the features learned are expected to be transformation-aware. Local feature consensus can boost the consistency among inlier matches and further enhance their discrimination against outliers. Moreover, inspecting the four outdoor scenes in the first four columns when matching the same landmark from a close shot to a remote one, existing approaches are sensitive to outlier correspondences since multiple pixels may be projected to the same location under the one-way mapping. Such an issue can be effectively tackled by the Siamese network with reciprocal constraints. Equipping with the two merits, our approaches successfully detect most of the inlier matches in all image pairs in Fig. \ref{fig:visual}, with the fewest outlier matches misidentified.

\subsection{Cross-view Pose Estimation}
Table \ref{tab:pose} shows the performances of cross-view pose estimation of different methods using mAP5$^\circ$ with or without RANSAC post-processing as done in \cite{dai2022ms2dg}. In viewing the performance gain over MSA-Net without RANSAC, our method obtains 18.24\% and 10.86\% for the known and unknown scenes of the YFCC100M dataset, and 41.91\% and 30.91\% for that of SUN3D dataset. Moreover, it surpasses all methods in all cases except for the known scene of the SUN3D dataset when RANSAC is applied. In principle, incorporating the reciprocal loss allows the network to predict the cross-view essential matrix and the reverse essential matrix of an image pair simultaneously, whereby the consistency between the two matrices is more strengthened. Meanwhile, training the proposed Siamese network with the reverse correspondence set actually doubles the training data for optimizing the underlying base network. All these factors facilitate the proposed Siamese network to be a more robust cross-view pose estimator. 
\begin{table}[htbp]
\small
  \centering
  \caption{Quantitative results of camera pose estimation on the YFCC100M and SUN3D datasets. The mAP5$^\circ$(\%) without/with RANSAC is shown.}
    \setlength{\tabcolsep}{1mm}{
    \begin{tabular}{c|cc|cc}
    \hline
    \multirow{2}{*}{Matcher} & \multicolumn{2}{c|}{YFCC100M(\%)} & \multicolumn{2}{c}{SUN3D(\%)} \\
\cline{2-5}          & Known  & Unknown  & Known  & Unknown  \\
    \hline
    Point-Net++ & 10.49/33.78 & 16.48/46.25 & 10.58/19.17 & 8.10/15.29 \\
    \hline
    DFE   & 19.13/36.46 & 30.27/51.16 & 14.05/21.32 & 12.06/16.26 \\
    \hline
    ACNet & 29.17/40.32 & 33.06/50.89 & 18.86/22.12 & 14.12/16.99 \\
    \hline
    CNe   & 13.81/34.55 & 23.95/48.03 & 11.55/20.60 & 9.30/16.40 \\
    \hline
    OA-Net++ & 32.57/41.53 & 38.95/52.59 & 20.86/22.31 & 16.18/17.18 \\
    \hline
    NM-Net & -/-   & 32.93/51.90 & -/-   & 14.13/16.86 \\
    \hline
    T-Net & 44.49/47.00 & 52.28/56.08 & 24.96/\textbf{23.81} & 19.71/18.00 \\
    \hline
    MS$^2$DG-Net & 38.36/45.34 & 49.13/57.68 & 22.20/23.00 & 17.84/17.79 \\
    \hline 			
    MSA-Net & 40.30/44.42 & 50.65/56.55 & 17.61/21.76 & 15.11/17.07 \\
    \hline 
    Ours(MSA-LFC) & 44.60/46.19 & 53.62/57.25 & 22.84/22.64 & 18.41/17.80 \\
    \hline
    Ours  & \textbf{47.65/47.23} & \textbf{56.15}/\textbf{58.67} & \textbf{24.99}/23.02 & \textbf{19.78}/\textbf{18.42} \\
    \hline
    \end{tabular}}%
  \label{tab:pose}%
\end{table}%

\subsection{Ablation Studies}
\label{subsec:ablation}
\begin{figure}[!htb]
\centering
\includegraphics[width=.48\textwidth]{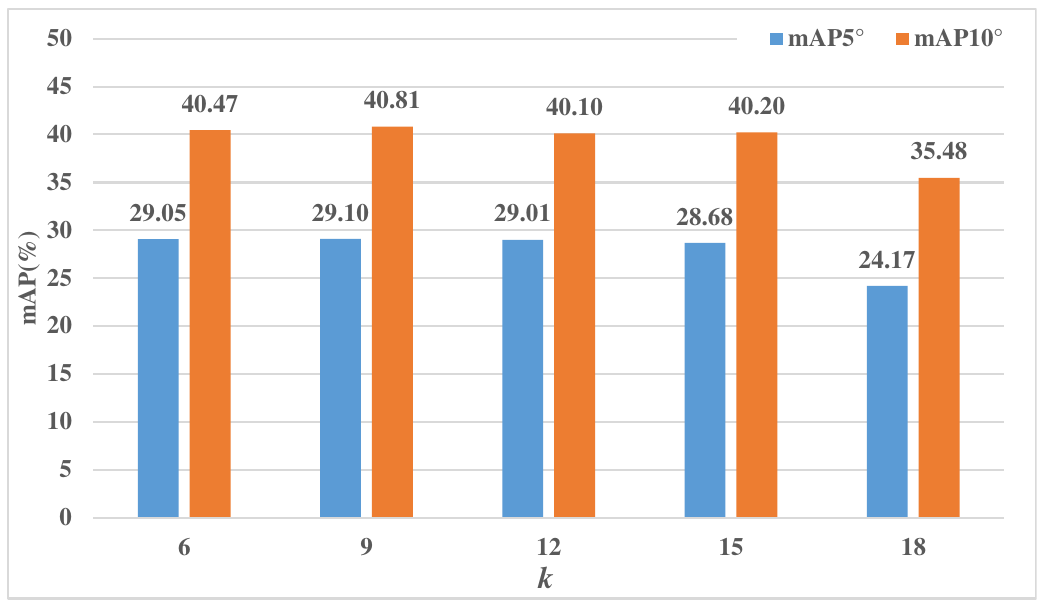}
\caption{Parametric analysis of $k$ on known scenes of YFCC100M.}
\label{fig:ktest}
\end{figure}

\noindent\textbf{Setting $k$}. 
The number of nearest neighbors $k$ determines the range of local consensus. 
To evaluate its impact efficiently, we randomly extract 1/5 image pairs from the training set of YFCC100M and train LFC block injected MSA-Net with different $k$. Results on the known scenes of YFCC100M are shown in Fig. \ref{fig:ktest}. It suggests the performance arises as $k$ increases in the beginning but degenerates after 9, probably due to that outliers gradually dominate the neighbor set of an inlier match and harm the effectiveness of feature consensus. As a large $k$ also incurs a more computational cost, we set $k$ = 9 in our experiments.

\noindent\textbf{Effectiveness of different components}. Table \ref{tab:ablation} shows the performance on the YFCC100M dataset when adding different components to MSA-Net. As shown, LFC injection (row 3) or Siamese extension (row 4) of MSA-Net alone can achieve performance promotion over the baseline method (row 1), which demonstrates the effectiveness of both components in the correspondence learning task.
Moreover, in the second row, we replace the deformable attention-based weighting with an MLP for feature fusion. Row 1 and 2 suggest that neighboring feature consensus can help to generate a more discriminative feature, while row 3 further justifies the usage of deformable attention-based weights. Finally,
integrating all components obtains compounded performance gain. 
\begin{table}[htbp]
  \centering
  \caption{Ablation studies on the YFCC100M dataset. mAP5$^\circ$(\%) without/with RANSAC are shown. LFC$_1$: Neighboring feature consensus. LFC$_2$: Deformable attention-based feature fusion.}
  \setlength{\tabcolsep}{1mm}{
    \begin{tabular}{cccc|cc}
    \hline
    MSA   & LFC$_1$  & LFC$_2$ & Siamese & Known & Unknown \\
    \hline
    \checkmark     &     &       &       & 40.30/44.42 & 50.65/56.55 \\
    \checkmark     &\checkmark     &       &       & 43.96/46.07 & 51.60/56.62 \\
    \checkmark     &\checkmark     & \checkmark     &       & 44.60/46.19& 53.62/57.25 \\
    \checkmark     &     &       & \checkmark     & 42.38/45.45 & 52.87/56.77 \\
    \checkmark     &\checkmark     & \checkmark     & \checkmark     & \textbf{47.65/47.23} & \textbf{56.15/58.67} \\
    \hline
    \end{tabular}%
    }
  \label{tab:ablation}
\end{table}%

\noindent\textbf{Design of Siamese network structure}. Table \ref{tab:Siamese} shows the performance of two Siamese structures presented in Fig. \ref{fig:siamese}. Overall, the network of Fig. \ref{fig:siamese}(b) achieves slightly better performance than that of Fig. \ref{fig:siamese}(a) in terms of most measures, while the latter design requires more computational cost. Therefore, the Siamese structure of Fig. \ref{fig:siamese}(b) is selected as the final design.
\begin{table}[htbp]
  \centering
  \caption{Performance comparison of two Siamese network designs presented in Fig. \ref{fig:siamese} on the YFCC100M dataset.}
    \small
    \setlength{\tabcolsep}{0.8mm}{
    \begin{tabular}{c|cccc|cccc}
    \hline
    \multirow{2}{*}{} & \multicolumn{4}{c|}{Known Scene} & \multicolumn{4}{c}{Unknown Scene} \\
    \cline{2-9}  &\multicolumn{1}{c}{mAP5$^\circ$} & \multicolumn{1}{c}{$P$} & \multicolumn{1}{c}{$R$} & \multicolumn{1}{c|}{$F$} & \multicolumn{1}{c}{mAP5$^\circ$} & \multicolumn{1}{c}{$P$} & \multicolumn{1}{c}{$R$} & \multicolumn{1}{c}{$F$} \\
    \hline
    (a) & 46.22 & 64.71 & \textbf{92.07} & 76.00  & 55.37 & 59.05 & \textbf{89.25} & 71.08 \\
    \hline
    (b) & \textbf{47.65} & \textbf{65.47} & 91.94 & \textbf{76.48 } & \textbf{56.15} & \textbf{60.84}  & 88.66 & \textbf{72.16} \\
    \hline
    \end{tabular}%
    }
  \label{tab:Siamese}%
\end{table}%

\section{Conclusions}
This paper introduces two techniques to boost the existing two-view correspondence learning framework. Firstly, a local feature consensus block is designed to augment neighboring features of a match with an attention-like mutual consensus, followed by attention-inspired deformable neighboring feature aggregation to reconstruct more discriminative correspondence features. Secondly, as existing approaches employ one-way projection supervision for network training, we propose to extend the network to a Siamese one and explore a reciprocal loss to better ensure bidirectional matching consistency. We apply the two proposals to MSA-Net, achieving state-of-the-art performance for the tasks of correspondence prediction and relative pose estimation on the YFCC100M and SUN3D datasets.

\begin{acks}
This work is supported by the Natural Science Foundation of Anhui Province (2108085MF210) and the Key Natural Science Fund of the Department of Education of Anhui Province (KJ2021A0042).
\end{acks}

\bibliographystyle{ACM-Reference-Format}
\bibliography{acmart}
\end{document}